\documentclass{article}

\usepackage[preprint]{neurips_2025}

\usepackage[utf8]{inputenc} 
\usepackage[T1]{fontenc}    
\usepackage{hyperref}       
\usepackage{url}            
\usepackage{booktabs}       
\usepackage{amsfonts}       
\usepackage{nicefrac}       
\usepackage{microtype}      

\usepackage{graphicx}
\usepackage{amsmath}  
\usepackage{amsfonts} 
\usepackage{multirow}
\usepackage{setspace} 
\usepackage[table]{xcolor}
\usepackage{amsmath, amssymb}
\usepackage{algorithm}
\usepackage{algpseudocode}
\usepackage{xcolor}
\usepackage{amssymb}
\usepackage{caption}
\usepackage{subcaption}
\usepackage{float}
\usepackage{geometry}
\title{DrSR: LLM based Scientific Equation Discovery with Dual Reasoning from Data and Experience}

\author{
  Runxiang Wang\textsuperscript{1,2} \quad 
  Boxiao Wang\textsuperscript{2,4} \quad 
  Kai Li\textsuperscript{2,3}\thanks{Corresponding authors.} \quad 
  Yifan Zhang\textsuperscript{1,2}\footnotemark[1] \quad 
  Jian Cheng\textsuperscript{1,2} \\
  \textsuperscript{1}School of Advanced Interdisciplinary Sciences, University of Chinese Academy of Sciences \\
  \textsuperscript{2}C\textsuperscript{2}DL, Institute of Automation, Chinese Academy of Sciences \\
 \textsuperscript{3} School of Artificial Intelligence, University of Chinese Academy of Sciences\\
  \textsuperscript{4}School of Mathematical Sciences, University of Chinese Academy of Sciences \\
}

\begin{document}

\maketitle

\begin{abstract}
Symbolic regression is a fundamental tool for discovering interpretable mathematical expressions from data, with broad applications across scientific and engineering domains. Recently, large language models (LLMs) have demonstrated strong performance in this task, leveraging embedded scientific priors and reasoning capabilities to surpass traditional methods. However, existing LLM-based approaches, such as LLM-SR, often over-rely on internal priors, lacking explicit data understanding and systematic reflection during equation generation. To address these limitations, we propose DrSR (Dual Reasoning Symbolic Regression), a framework that combines data-driven insight with reflective learning to enhance both robustness and discovery capability. Specifically, DrSR guides LLMs to analyze structural relationships—e.g., monotonicity, nonlinearity, and correlation—within the data to generate structured descriptions. Simultaneously, it monitors equation performance and establishes a feedback loop to refine subsequent generations. By integrating data understanding and generation reflection in a closed loop, DrSR enables more efficient exploration of the symbolic expression space. Experiments across interdisciplinary datasets in physics, chemistry, biology, and materials science demonstrate that DrSR substantially improves the valid equation rate and consistently outperforms both classical and recent LLM-based methods in terms of accuracy, generalization, and search efficiency—underscoring its potential for scientific equation discovery.
\end{abstract}

\section{Introduction}
\label{headings}

Symbolic regression (SR)~\cite{makke2024interpretable} aims to recover interpretable mathematical expressions from observational data and plays a critical role in scientific modeling—ranging from uncovering physical laws~\cite{lemos2023rediscovering,davis2023discovery}, to modeling chemical kinetics~\cite{batra2021emerging,hernandez2019fast}, to understanding financial dynamics~\cite{yu2023generating,xu2024hrft}. The resulting equations support extrapolation, cross-domain transfer, and mechanistic understanding~\cite{schmidt2009distilling}.

The core challenge in SR lies in navigating the vast, combinatorial space of mathematical expressions to identify models that both fit the data and remain interpretable. To address this, a range of algorithmic strategies have been proposed. Classical approaches based on genetic programming (GP)~\cite{schmidt2009distilling,cranmer2023interpretable} evolve expression trees via mutation and crossover. Reinforcement learning (RL) methods~\cite{petersen2019deep} treat symbolic equation discovery as a sequential decision-making problem, using policy gradients to explore the space of symbolic programs. Transformer-based neural models~\cite{biggio2021neural,kamienny2022end} have been employed to directly generate symbolic expressions by modeling equations as sequences.

\begin{figure}[htbp]
  \centering
  \includegraphics[width=0.945\linewidth]{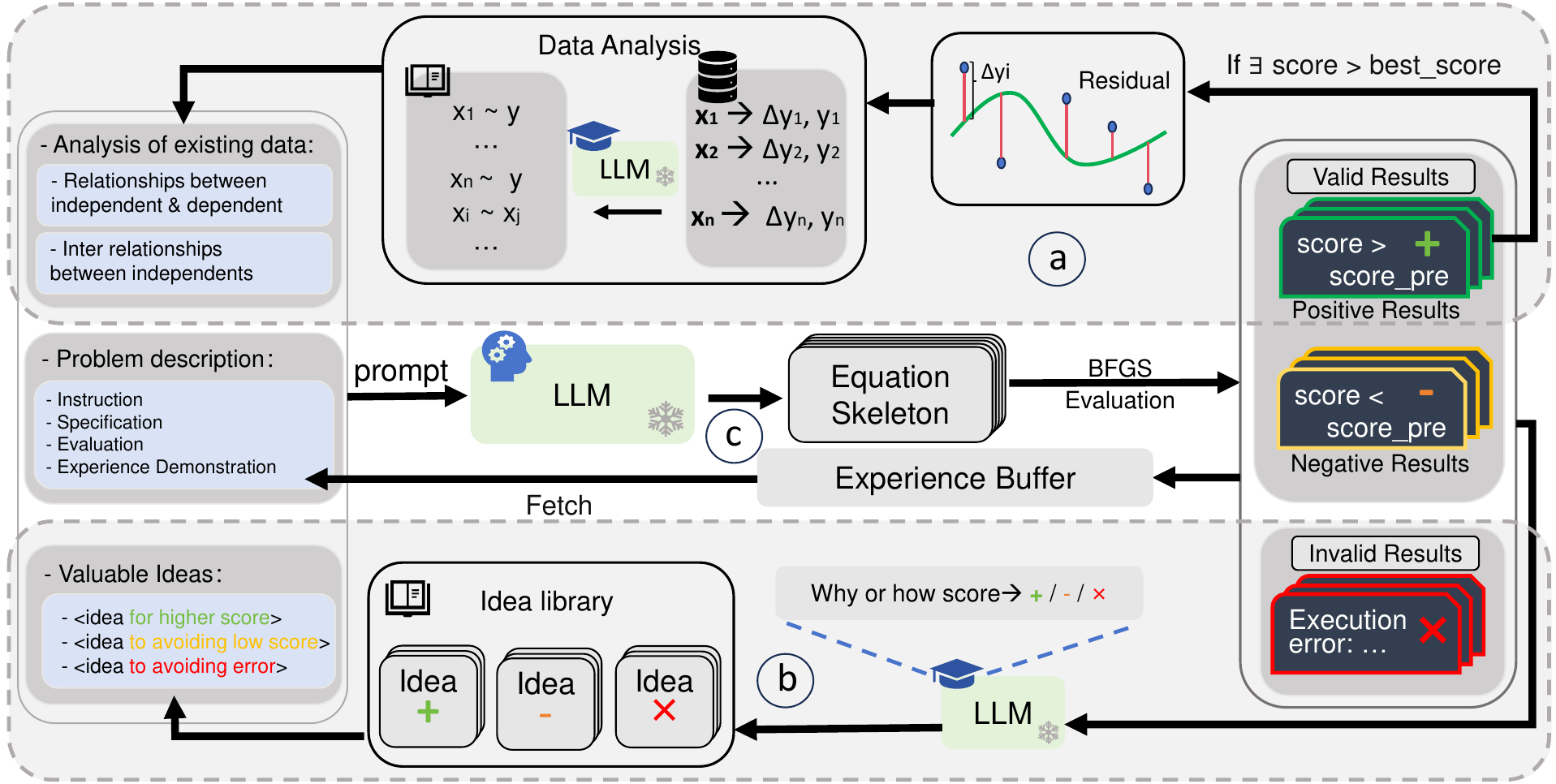}
  \caption{
  \textbf{Overview of the DrSR framework.}
  (a) \textit{Data-aware Insight:} the LLM analyzes variable relationships to derive structured, data-informed insights; 
  (b) \textit{Inductive Idea Extraction:} based on equation fitness, the LLM summarizes successful and failed patterns into an evolving idea library;
  (c) \textit{Equation Generation and Selection:} guided by (a) and (b), the LLM generates new equation skeletons, optimizes parameters, and caches the best-performing results to inform future generations.
  }
  \label{fig:drsr_framework}
\end{figure}

More recently, LLMs have further advanced this area by generating candidate equations from textual prompts. For example, LLM-SR~\cite{shojaee2024llm} proposes new expressions conditioned on task descriptions and prior candidates, combining LLM generation with numeric optimization. 
However, such methods overly rely on the scientific priors embedded within LLMs while neglecting structural insights into variable relationships—effectively ignoring direct observation and analysis of the raw data. This limits the accuracy and generalizability of the generated equations. Moreover, the lack of systematic reflection on generation history often leads to the repeated production of invalid expressions (e.g., syntax errors, numerical overflows), reducing both the efficiency and stability of SR. To address these issues, a more effective SR framework is needed—one that not only incorporates scientific priors flexibly, but also analyzes data structures and learns from past generation behavior. Such an approach would better align with the scientist’s cognitive process: inducing patterns from observations, reflecting through trial and error, and efficiently navigating the vast hypothesis space to uncover interpretable and predictive models.

To this end, we propose DrSR (Dual Reasoning Symbolic Regression) (Fig \ref{fig:drsr_framework}), a framework that augments LLM-based generation with two synergistic reasoning mechanisms: \emph{data-aware insight}, which derives structural patterns from raw data, and \emph{inductive idea extraction}, which reflects on generation outcomes to distill reusable strategies. Specifically, building upon the generation-evaluation loop of LLM-SR, a data-aware module prompts the LLM to extract structural patterns from a sampled subset of raw data, providing task-specific priors on variable relationships. As better-fitting equations are discovered, these insights are incrementally updated (Fig~\ref{fig:drsr_framework}(a)). Secondly, DrSR introduces an experience-based module that categorizes newly generated equations based on evaluation outcomes and prompts the model to reflect on success or failure. The resulting strategies—summarized as structured ``ideas''—are stored in a dynamic idea library and reused to improve future equation generation (Fig~\ref{fig:drsr_framework}(b)). By combining data-driven interpretation with behavior-level feedback, DrSR enables continual refinement of both symbolic priors and generative heuristics, resulting in more efficient and reliable symbolic equation discovery.

We evaluate DrSR on six benchmarks spanning physics, chemistry, biology, and materials science, using Mixtral-8x7B-Instruct-v0.1~\cite{jiang2024mixtral} and LLaMA3.1-8B-Instruct~\cite{kassianik2025llama} as backbones. Compared to traditional SR methods and LLM-based baselines, DrSR achieves state-of-the-art performance across multiple metrics, including accuracy, generalization, valid expression rate, and convergence speed.

\section{Preliminaries}
\label{headings}

In SR, the objective is to discover a compact symbolic expression \( f(\mathbf{x}) \) that approximates an unknown mapping \( f^{real} : \mathbb{R}^d \rightarrow \mathbb{R} \) based on a dataset of input-output pairs \( D = \{(\mathbf{x}_i, y_i)\}_{i=1}^n \). SR methods aim to recover underlying mathematical relationships such that \( f(\mathbf{x}_i) \approx y_i \) for all \( i \). The discovered expression is expected to not only fit the observed data accurately but also generalize well to unseen inputs while maintaining interpretability. Model performance is typically evaluated via the negative mean squared error:
\(
\text{Score}(f, D) = -\frac{1}{n} \sum_{i=1}^{n} (f(\mathbf{x}_i) - y_i)^2.
\)

Our work builds upon LLM-SR~\cite{shojaee2024llm}, a recent framework that utilizes LLM to generate interpretable equations from data. To contextualize our proposed DrSR, we briefly review the core workflow of LLM-SR. LLM-SR discovers equations by combining the scientific knowledge embedded in LLMs with their powerful code generation capabilities, prompting the LLM to generate equation skeletons \( \mathcal{F} \) with learnable parameters. The overall procedure consists of three key components:

\textbf{Hypothesis Generation.} The LLM receives a structured prompt comprising: a task description \( \mathcal{T} \), variable constraints, evaluation criteria, and reference equations. Based on this, it generates equation skeletons aligned with the physical meaning of variables.

\textbf{Hypothesis Evaluation.} Each skeleton is fitted to data by optimizing its parameters. The resulting function is scored via the fitness metric \( \text{Score}(f, D) \).

\textbf{Experience Management.} A memory buffer stores high-scoring equations, which are reused as demonstrations in future prompts to improve generation quality.

While LLM-SR effectively leverages prior knowledge embedded in LLMs, it suffers from two critical drawbacks: First, its understanding of data remains shallow: although LLM-SR can generate equation skeletons via language prompts and refine them through parameter optimization, it lacks direct engagement with the data itself. As a result, it fails to capture complex inter-variable relationships that are crucial for SR, limiting the reasoning capacity of LLMs. Second, it lacks a reflective mechanism to evaluate and learn from its own behavior. Without such self-correction, LLM-SR frequently produces invalid expressions—such as syntactically incorrect forms, numerical instabilities, or inconsistent variable usage—ultimately degrading performance and robustness.

\section{Method}
\label{headings}

To overcome these limitations, we propose DrSR (Dual Reasoning Symbolic Regression), a novel framework that enhances SR by equipping LLM with two synergistic reasoning pathways: \textit{Data-aware Insight} and \textit{Inductive Idea Extraction}. While retaining the generative strength of LLM-SR in leveraging scientific priors, DrSR introduces a closed-loop reasoning cycle that enables the model to both analyze raw data and reflect on its generation history. DrSR transforms SR into a dynamic cognitive process: observing to understand and reflecting to improve.

To implement this, DrSR instantiates three role-specific, LLM-based modules: \(\pi_{\text{data}}\) for data-aware insight extraction, \(\pi_{\text{idea}}\) for modeling experience summarization, and \(\pi_{\text{main}}\) for equation generation. All modules share the same LLM backbone.

\subsection{Data-aware Insight}
\label{sec:data_insight}

To enhance the structural awareness of SR, we introduce a dedicated \textit{Data-Analysis-LLM} $\pi_{\text{data}}$, which analyzes input-output pairs and residuals to generate structured, iteratively updated insights $\mathcal{D}$. These insights encode variable relationships and serve as evolving structural priors for the main model $\pi_{\text{main}}$ to guide equation generation.

\subsubsection{Initial Data-aware Insight}
To control input length and preserve key behavioral patterns, we uniformly sample 100 input-output pairs from the original dataset $D = \{(\mathbf{x}_i, y_i)\}_{i=1}^n$ to form $D_0^{\text{resample}}$. Feeding this into $\pi_{\text{data}}$ yields the initial insight $\mathcal{D}_0$, capturing basic patterns such as monotonicity, nonlinearity, and variable correlation.

\subsubsection{Iterative Residual-based Refinement}
At each iteration $t$, if a candidate $f^*$ improves upon the current best score $s^{*}$, $\pi_{\text{data}}$ produces an updated insight $\mathcal{D}_{\text{new}}$ to refine the future generations of $\pi_{\text{main}}$.

\textbf{Prompt Construction.}
Residuals provide informative signals about parts of the data that are poorly captured by the current model, often revealing latent variable interactions or unmodeled structures. To help $\pi_{\text{data}}$ focus on these regions, we compute residuals as $res_{t,i} = y_i - f^*(\mathbf{x}_i)$ and construct an augmented dataset $D_t = \{(\mathbf{x}_i, y_i, res_{t,i})\}_{i=1}^n$. We then uniformly resample 100 examples from $D_t$ to form $D_t^{\text{resample}}$, which is used to generate refined structural insights in the next iteration.

\textbf{Insight Generation.}
Using $D_t^{\text{resample}}$, the candidate $f^*$, and prior insight $\mathcal{D}_{\text{pre}}$, $\pi_{\text{data}}$ generates a refined interpretation. This includes local monotonicity changes, nonlinear interactions (e.g., products, ratios), and more specific variable correlations, forming a stronger prior for guiding generation.

As iterations proceed, insights evolve from coarse global patterns to finer local and higher-order structures, guiding $\pi_{\text{main}}$'s generation process toward more precise, interpretable expressions.

\subsection{Inductive Idea Extraction}
\label{sec:idea_extraction}

To improve search efficiency and robustness, we introduce an \textit{Idea-Extraction-LLM} $\pi_{\text{idea}}$, which mimics scientific reflection by extracting actionable ideas from model feedback. At each iteration, it analyzes the generated equations and summarizes both successes and failures into structured heuristics. These ideas encode generation strategies, common pitfalls, or error-avoidance techniques, enabling the model to iteratively refine its behavior and improve symbolic discovery.

\subsubsection{Extraction Process}

After $\pi_{\text{main}}$ generates and optimizes equations, each evaluated candidate $f$ is categorized as: \textbf{1) Positive.} Outperforms the reference, suggesting effective modeling. \textbf{2) Negative.} Performs worse than the reference, indicating potential flaws. \textbf{3) Invalid.} Cannot be evaluated due to syntax errors, numerical faults, or variable mismatches.

For each type, $\pi_{\text{idea}}$ is prompted with tailored templates (see Appendix C) to elicit structural heuristics: e.g., strengths of winning patterns, causes of performance drops, or failure modes and remedies. The extracted ideas are categorized and stored for downstream use.

\begin{minipage}{0.445\textwidth}
\vspace{-0.16em}
  \captionsetup{type=algorithm}
  \begin{algorithm}[H]
    \caption{DrSR}
    \label{alg:drsr}
    \noindent\textbf{Input:}  LLM $\pi_{data}$, 
     LLM $\pi_{idea}$, 
     LLM $\pi_{main}$, 
    dataset ${D}$, 
    $T$ iterations, $k$ in-context examples, 
    $b$ samples per iteration,
    proportion of recent ideas $\lambda$ 

    \begin{algorithmic}
    \small

    \State $P_0, \mathcal{D}, \mathcal{L}, \mathcal{P} \gets \textsc{InitAll}()$
    \State $f^*, s^* \gets \texttt{null}, -\infty$

    \For{$t = 1$ to $T$}

        \State $\mathcal{F}_t \gets \textsc{SampleLLM}(\pi_{main}, \mathcal{P}, k, b)$

        \For{$f \in \mathcal{F}_t$}
            \State $s \gets \text{Score}(f, {D})$

            \State $\mathcal{L} \gets$ \textsc{IdeaSummarization}$(f, \mathcal{P},$
            \State \hspace{\algorithmicindent} $\pi_{idea}, \mathcal{D}, s^*, \mathcal{L})$

            \If{$s > s^*$}
                \State $\mathcal{D} \gets$ \textsc{DataAnalysis}$(f, \mathcal{D},$
                \State \hspace{\algorithmicindent} $D, \pi_{data})$
                \State $f^*, s^* \gets f, s$
                \State $P_t \gets P_{t-1} \cup \{(f, s)\}$
            \EndIf
        \EndFor
        \State \textcolor{gray}{{\scriptsize\#} Sample from latest $\lambda$ proportion}
        \State $\mathcal{L}_{\text{sampled}} \gets \textsc{SampleRecentIdeas}(\mathcal{L}, \lambda)$
        \State $\mathcal{P} \gets \textsc{UpdatePrompt}(\mathcal{L}_{\text{sampled}}, \mathcal{D})$
    \EndFor
    \end{algorithmic}

    \noindent\textbf{Output:} $f^*, s^*$

  \end{algorithm}
\end{minipage}
  \hfill
\begin{minipage}{0.54\textwidth}
  \captionsetup{type=algorithm}
  \begin{algorithm}[H]
    \caption{\textsc{Idea-Summarization}}
    \label{alg:idea_summarization}
    \noindent\textbf{Input:} Equation $f$, 
    prompt history $\mathcal{P}$, 
    LLM $\pi_{idea}$, 
    dataset $D$, 
    best score $s^*$,
    Library $\mathcal{L}$
    
    \vspace{0.3em}

    \begin{algorithmic}
    \small

    \State result $\gets$ \textsc{EvaluateEquation}$(f, D)$

    \If{result contains \texttt{error}}
        \State category $\gets$ \texttt{INVALID}
    \ElsIf{$\text{Score}(f, D) > s^*$}
        \State category $\gets$ \texttt{POSITIVE}
    \Else
        \State category $\gets$ \texttt{NEGATIVE}
    \EndIf

    \State $\mathcal{I} \gets$ \textsc{IdeaExtraction}$(\mathcal{P}, f,
    \text{result}, \texttt{category}, \pi_{idea})$
    \State $\mathcal{L} \gets$ \textsc{IdeaLibraryUpdate}$(\mathcal{L},
    \mathcal{I}, \texttt{category})$
    \end{algorithmic}

    \noindent\textbf{Output:} $\mathcal{L}$

  \end{algorithm}

  \vspace{-2.3em}

  \begin{algorithm}[H]
    \caption{\textsc{Data-Analysis}}
    \label{alg:data_analysis}
    \noindent\textbf{Input:} Equation $f$, 
    Dataset $D$, 
    Previous insight $\mathcal{D}_{pre}$, 
    LLM $\pi_{data}$

    \begin{algorithmic}
    \small

    \State $\hat{y}_i = f(x_i; \theta^*)$, $res_{t,i} = y_i - \hat{y}_i$
    \State $D_t = \{(x_i, y_i, res_{t,i})\}_{i=1}^n$
    \State \textcolor{gray}{{\scriptsize\#} Resample: $(x_i', y_i', res_{t,i}') \sim_{\text{i.i.d.}} \text{Unif}(D_t)$}
    \State $D_t^{\text{resample}} = \{(x_i', y_i', res_{t,i}')\}_{i=1}^{n'}$
    \State $\mathcal{D}_{new} \gets \pi_{data}(D_t^{\text{resample}}, f, \mathcal{D}_{pre})$
    \end{algorithmic}

    \noindent\textbf{Output:} $\mathcal{D}_{new}$

  \end{algorithm}
\end{minipage}

\subsubsection{Idea Library}
Extracted ideas are organized in the Idea Library $\mathcal{L}$ as follows:

\textbf{Positive Ideas:} Capture patterns from successful equations, such as the inclusion of high-order or domain-specific terms (e.g., sinusoidal or multiplicative components).

\textbf{Negative Ideas:} Highlight structural flaws—e.g., overcomplexity or mismatches with data characteristics—helping to steer the model away from ineffective regions.

\textbf{Error Avoidance Ideas:} Focus on eliminating invalid outputs by encoding rules that ensure syntactic and semantic correctness.

All ideas are stored in JSON format with metadata including context and fitness score. In each iteration, recently extracted ideas are sampled and injected into prompts of \(\pi_{\text{main}}\), supporting continual adaptation and improved generation quality.

\subsection{Overall Workflow of DrSR}

DrSR integrates two complementary reasoning pathways—\emph{data-driven insight} and \emph{experience-based reflection}—to enable LLMs to perform SR with self-correcting and structurally informed capabilities. In each iteration, the model not only generates and evaluates candidate equations but also refines its understanding of the data and summarizes patterns from prior attempts. This yields a closed-loop optimization process that improves both equation quality and search efficiency over time. The pseudocode of DrSR is detailed in Algorithm~\ref{alg:drsr}.

\subsection{A Bayesian Perspective on DrSR}

The superior performance of DrSR arises not merely from architectural enhancements, but from a principled rethinking of how LLMs conduct SR. We reinterpret its generative process through the lens of Bayesian modeling to highlight its strengths over prior approaches such as LLM-SR.

In SR, the goal is to identify an expression \( f \) from dataset \( D = \left\{ (\mathbf{x}_i, y_i) \right\}_{i=1}^n \) that is both predictive and interpretable. Following the Bayesian formulation~\cite{merler2024context}, LLM-SR can be viewed as maximizing:
\begin{equation}
\arg\max_{f} p_{\text{LLM}}(f \mid D) = \arg\max_{f} p_{\text{LLM}}(D \mid f) \cdot p_{\text{LLM}}(f),
\end{equation}
where \( p_{\text{LLM}}(f) \) captures the prior belief of LLMs about the plausibility of expression \( f \), reflecting how likely the LLM is to generate \( f \) based on its learned distribution over symbolic forms, and \( p_{\text{LLM}}(D \mid f) \) measures the explanatory adequacy of the expression \( f \) with respect to the dataset \( D \).
However, this approach relies heavily on language-induced priors and lacks two key elements: structural understanding of the data and reflective learning from prior generations.

DrSR addresses these limitations by introducing two adaptive cognitive variables: the data insight \( \mathcal{D} \), summarizing variable relationships, and the idea library \( \mathcal{I} \), capturing learned generation strategies. With these, the objective becomes:
\begin{equation}
\arg\max_{f, \mathcal{D}, \mathcal{I}} p_{\text{LLM}}(f, \mathcal{D}, \mathcal{I} \mid D) 
= \arg\max_{f, \mathcal{D}, \mathcal{I}} p_{\text{LLM}}(D \mid f) \cdot p_{\text{LLM}}(f \mid \mathcal{D}, \mathcal{I}) \cdot p_{\text{LLM}}(\mathcal{D}, \mathcal{I}),
\end{equation}
where \( p_{\text{LLM}}(f \mid \mathcal{D}, \mathcal{I}) \) serves as a cognitively enriched prior that guides equation generation, and \( p_{\text{LLM}}(\mathcal{D}, \mathcal{I}) \) models the evolution of structural knowledge over time.

This formulation does not change the task objective but enhances the generative prior from a static, language-based prior \( p_{\text{LLM}}(f) \) to a dynamic, context-aware form conditioned on evolving insight and experience. As \( \mathcal{D} \) and \( \mathcal{I} \) are iteratively refined, the structural preferences of the model become sharper, enabling more efficient and stable exploration of the symbolic expression space.

\section{Experiments}
\label{headings}

\subsection{Benchmarks and Datasets}

We evaluate DrSR on six representative SR datasets spanning physics, biology, chemistry, and materials science. Four datasets are taken from LLM-SR~\cite{shojaee2024llm}, and two from the LLM-SRBench suite~\cite{shojaee2025llm}, including LSR-Transform and LSR-Synth. Full details are available in Appendix B.

\textbf{Nonlinear Oscillators.}  
This dataset simulates two nonlinear damped oscillator systems, each governed by second-order differential equations involving displacement, velocity, and external forcing. Compared to classical oscillators, they incorporate cubic, multiplicative, and non-polynomial terms, yielding complex variable dependencies and heightened modeling difficulty.

\textbf{Bacterial Growth.} 
This dataset models the growth rate of E. coli under varying conditions, incorporating population density, substrate concentration, temperature, and pH. The ground-truth equation involves nonlinear effects of temperature and pH, reflecting complex interactions among variables. The task is to recover expressions that capture the dynamic behavior of bacterial proliferation~\cite{Monod1949The, rosso1995convenient}.

\textbf{Stress--Strain Behavior.} 
This dataset captures the mechanical response of 6061-T651 aluminum alloy under varying strain and temperature, based on tensile tests conducted across six temperatures (20$^\circ$C to 300$^\circ$C). It serves as a realistic benchmark for modeling stress as a function of strain and thermal conditions, and is widely applicable in structural engineering and materials science.

\textbf{LSR-Transform.} 
This dataset, from LLM-SRBench, tests the ability of LLMs to recover structurally altered expressions. The target functions originate from Feynman equations~\cite{udrescu2020ai}, transformed via variable substitutions and symbolic rewrites to increase complexity. Due to computational constraints, we select two representative subsets---\texttt{I.37.4\_0\_1} and \texttt{III.4.33\_3\_0}---for experimentation.

\textbf{LSR-Synth.} 
This synthetic benchmark from LLM-SRBench~\cite{shojaee2025llm} includes 129 SR tasks constructed by mixing canonical scientific terms with synthetic expressions. The dataset spans four domains—chemistry, physics, biology, and materials science—and aims to test LLMs’ ability to generalize across hybrid symbolic forms. Due to domain overlap with~\cite{shojaee2024llm} and compute limits, we focus on the chemistry subset \texttt{CRK0}.

\subsection{Baselines}

We compare DrSR with six representative SR methods, spanning classical, deep learning, and LLM-based paradigms: gplearn, PySR~\cite{cranmer2023interpretable}, DSR~\cite{petersen2019deep}, uDSR~\cite{landajuela2022unified}, LaSR~\cite{grayeli2024symbolic}, and LLM-SR~\cite{shojaee2024llm}.

gplearn uses GP to represent symbolic expressions as trees and explores the hypothesis space via mutation and crossover. PySR extends GP with parallelization and heuristics like simulated annealing and adaptive parsimony to improve efficiency. DSR employs RL to train neural networks that generate symbolic formulas. uDSR enhances DSR by combining neural-guided GP with large-scale pretraining. LaSR integrates LLMs with GP, using zero-shot prompts to guide the evolution of abstract equation representations. LLM-SR treats equations as programs, generating equation skeletons with LLMs and refining them via evolutionary search. These baselines represent diverse SR approaches and serve as strong comparators to evaluate the effectiveness of DrSR.

\subsection{Evaluation Metrics}

We evaluate SR models using two complementary metrics: Accuracy under error tolerance (\(\text{ACC}_\tau\)) and Normalized Mean Squared Error (NMSE), which respectively measure tolerance-aware generalization and numeric precision. Specifically \(\text{ACC}_\tau\) is defined as the proportion of test points with error below a threshold \(\tau\): \( \text{ACC}_\tau = \frac{1}{N_{\text{test}}} \sum_{i=1}^{N_{\text{test}}} \mathbf{1} \left( \left| \frac{f(x_i) - y_i}{y_i} \right| \leq \tau \right), \) where \(\mathbf{1}(\cdot)\) is the indicator function. NMSE quantifies prediction error normalized by target variance: \( \text{NMSE} = \frac{\sum_{i=1}^{N_{\text{test}}} (f(x_i) - y_i)^2}{\sum_{i=1}^{N_{\text{test}}} (y_i - \bar{y})^2}, \) where \(\bar{y}\) is the mean of the true values. NMSE places more weight on large deviations, making it suitable for evaluating numeric precision. \(\text{ACC}_\tau\) captures ``good enough'' performance under bounded error, aligning with real-world tolerance requirements. Together, these metrics provide a holistic view of model performance, balancing exactness and practicality.

\subsection{DrSR Configuration}

For fair comparison, we use the same backbone LLMs (Mixtral-8x7B-Instruct-v0.1 and LLAMA3.1-8B-Instruct) across all LLM-based methods. In each iteration, $4$ candidate equation skeletons are sampled, and their parameters are optimized via BFGS. DrSR is evaluated with a maximum of $1000$ iterations per dataset, while baselines like LLM-SR, LaSR, and classical methods are allowed $2000$ iterations or more to ensure convergence. Further implementation details, including prompts and sampling settings, are provided in Appendices A and C.

\section{Findings}
\label{headings}

\subsection{Overall Performance Comparison}

\vspace{-1mm}

\begin{table}[t]
    \caption{Overall performance of DrSR and baseline methods on six symbolic regression benchmarks.}
    \centering
    \setstretch{1.5} 
    \label{tab:comparison}
    \resizebox{\textwidth}{!}{ 
    \large 
    \begin{tabular}{lcccccccccccc}
    \hline
    \multirow{2}{*}{\textbf{Model}} & \multicolumn{2}{c}{\textbf{Oscillation 1}} & \multicolumn{2}{c}{\textbf{Oscillation 2}} & \multicolumn{2}{c}{\textbf{E. coli growth}} & \multicolumn{2}{c}{\textbf{Stress-Strain}} & \multicolumn{2}{c}{\textbf{LSR-Transform-2Avg}} & \multicolumn{2}{c}{\textbf{LSR-Synth-CRK0}} \\
    \cline{2-13}
     & \textbf{Acc\textsubscript{0.001}(\%)↑} & \textbf{NMSE↓} & \textbf{Acc\textsubscript{0.001}(\%)↑} & \textbf{NMSE↓} & \textbf{Acc\textsubscript{0.1}(\%)↑} & \textbf{NMSE↓} & \textbf{Acc\textsubscript{0.1}(\%)↑} & \textbf{NMSE↓} & \textbf{Acc\textsubscript{0.1}(\%)↑} & \textbf{NMSE↓} & \textbf{Acc\textsubscript{0.1}(\%)↑} & \textbf{NMSE↓} \\
    \hline
    GPlern & 0.11 & 0.0972 & 0.05 & 0.2000 & 0.76 & 1.002 & 28.43 & 0.3496 & 86.67 & 0.0070 & 0.80 & 1.0373 \\
    PySR & 3.80 & 0.0003 & 7.02 & 0.0002 & 2.80 & 0.4068 & 70.60 & 0.0347 & 50.57 & 0.0034 & \cellcolor{gray!20}\textbf{100} & 8.49e-7 \\
    DSR  & 0.42 & 0.01916 & 0.24 & 0.1793 & 0.64 & 0.8294 & 59.85 & 0.3435 & 70.94 & 0.0150 & 2.00 & 1.32e-5 \\
    uDSR & 1.78 & 0.0002 & 0.36 & 0.0856 & 1.12 & 0.5059 & 59.15 & 0.0639 & 82.62 & 0.0012 & 98.60 & 1.44e-7 \\
    \hline
    LLM-SR (Mixtral) & 5.9 & 0.0001 & 7.62 & 4.59e-5 & 2.08 & 0.2282 & 68.10 & 0.0530 & 90.79 & 0.0036 & 77.60 & 1.09e-6 \\
    LLM-SR (Llama3.1) & 12.67 & 2.55e-5 & 8.2 & 4.70e-5 & 1.36 & 0.5815 & 76.21 & 0.0333 & 93.73 & 0.0008 & 87.00 & 2.52e-6 \\
    LaSR (Mixtral) & 1.99 & 0.3332 & 1.64 & 0.0058 & 2.32 & 0.2955 & 20.52 & 1.1923 & 91.94 & 0.0029 & 94.20 & 2.49e-6 \\
    LaSR (Llama-3.1) & 2.79 & 0.7485 & 1.09 & 0.0310 & 3.44 & 0.1349 & 71.84 & 0.0320 & 92.30 & 0.0031 & 95.40 & 6.52e-8 \\
    \hline
    DrSR (Mixtral) & \cellcolor{gray!20}\textbf{83.92} & \cellcolor{gray!20}\textbf{3.14e-7} & \cellcolor{gray!20}\textbf{99.94} & \cellcolor{gray!20}\textbf{1.80e-12} & \cellcolor{gray!20}\textbf{5.12} & \cellcolor{gray!20}\textbf{0.0195} & \cellcolor{gray!20}\textbf{88.28} & \cellcolor{gray!20}\textbf{0.0156} & 92.51 & 0.0055 & 95.20 & 8.87e-8 \\
    DrSR (Llama-3.1) & 77.98 & 5.40e-7 & 99.33 & 1.23e-9 & 3.64 & 0.0797 & 72.95 & 0.0230 & \cellcolor{gray!20}\textbf{96.32} & \cellcolor{gray!20}\textbf{0.0006} & 98.00 & \cellcolor{gray!20}\textbf{1.75e-8} \\
    \hline
    \end{tabular}
    }
    
\end{table}

As shown in Table~\ref{tab:comparison}, DrSR consistently achieves significantly lower NMSE and higher \(\text{Acc}_\tau\) across most datasets, outperforming both classical and LLM-based baselines. For example, on Oscillator 2 , DrSR (Mixtral) achieves an NMSE of \(1.80 \times 10^{-12}\), far surpassing LLM-SR (Mixtral)’s \(4.59 \times 10^{-5}\). In terms of accuracy, DrSR exceeds $90\%$ on several benchmarks with both Mixtral and LLaMA backbones—highlighting strong alignment with ground truth under strict error thresholds.

Traditional methods such as PySR show reasonable performance on some tasks, such as LSR-Synth-CRK0, but struggle on others. Meanwhile, LLM-SR and LaSR, though capable of generating syntactically valid expressions, often suffer from instability and limited data sensitivity—failing to leverage feedback for continuous improvement.

These results underscore the core advantage of DrSR: its dual reasoning architecture. The \textit{Data-aware Insight} module provides structural priors from inter-variable analysis, while the \textit{Idea Extraction} module captures and reuses generation heuristics. Together, they enable more guided, interpretable, and efficient exploration of the expression space.

\subsection{Generalization Evaluation}

\begin{figure}[htbp]
  \centering
  \includegraphics[width=0.9\textwidth]{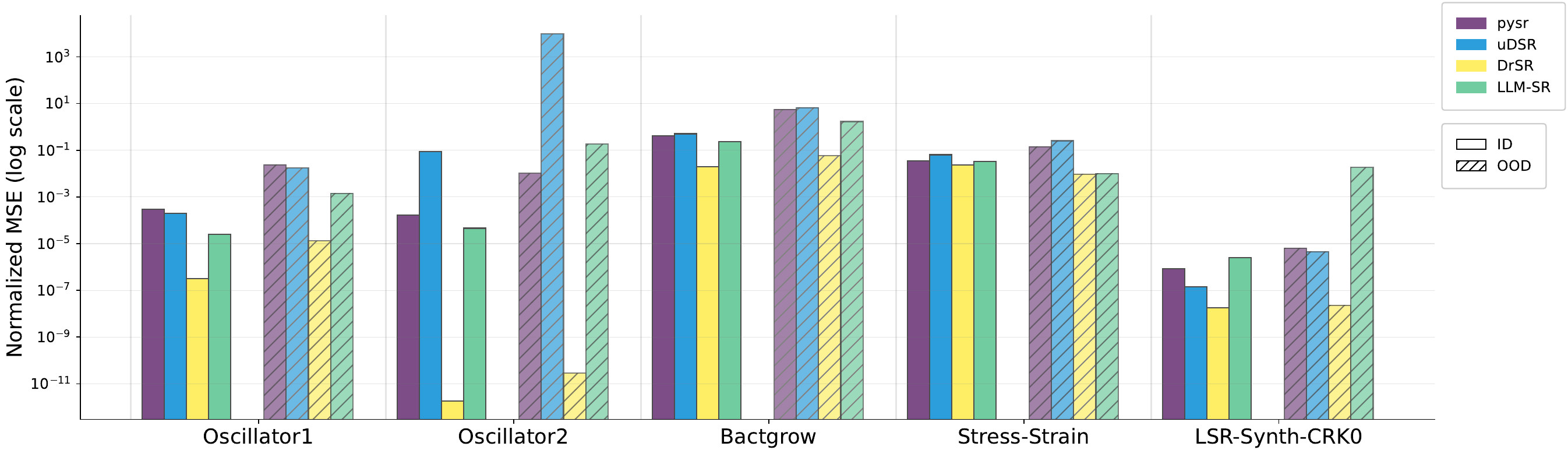}
  \caption{
  \textbf{Generalization across scientific domains under ID and OOD settings.}
  DrSR achieves the lowest NMSE across all tasks and settings, indicating strong generalization performance.
  }
  \label{fig:generalization}
\end{figure}

We further assess generalization by evaluating performance under both in-distribution (ID) and out-of-distribution (OOD) settings. Figure~\ref{fig:generalization} presents the NMSE of PySR, uDSR, LLM-SR, and DrSR under both conditions.
DrSR consistently achieves the best performance in both ID and OOD cases, confirming its robustness. Notably, on Oscillator 2, DrSR reduces OOD NMSE to \(2.80 \times 10^{-11}\), whereas uDSR degrades to \(9.59 \times 10^{3}\), indicating severe overfitting. 

DrSR also exhibits minimal performance drop between ID and OOD, suggesting it learns latent rules behind data rather than overfitting to distribution-specific patterns. This generalization is critical for real-world scientific modeling, where extrapolation beyond observed data is often required. Taken together, these findings position DrSR as a robust and accurate SR framework, capable of producing generalizable, interpretable models across a wide range of scientific domains.

\subsection{Training Efficiency Comparison}

\begin{figure}[htbp]
  \centering
  \includegraphics[width=0.9\textwidth]{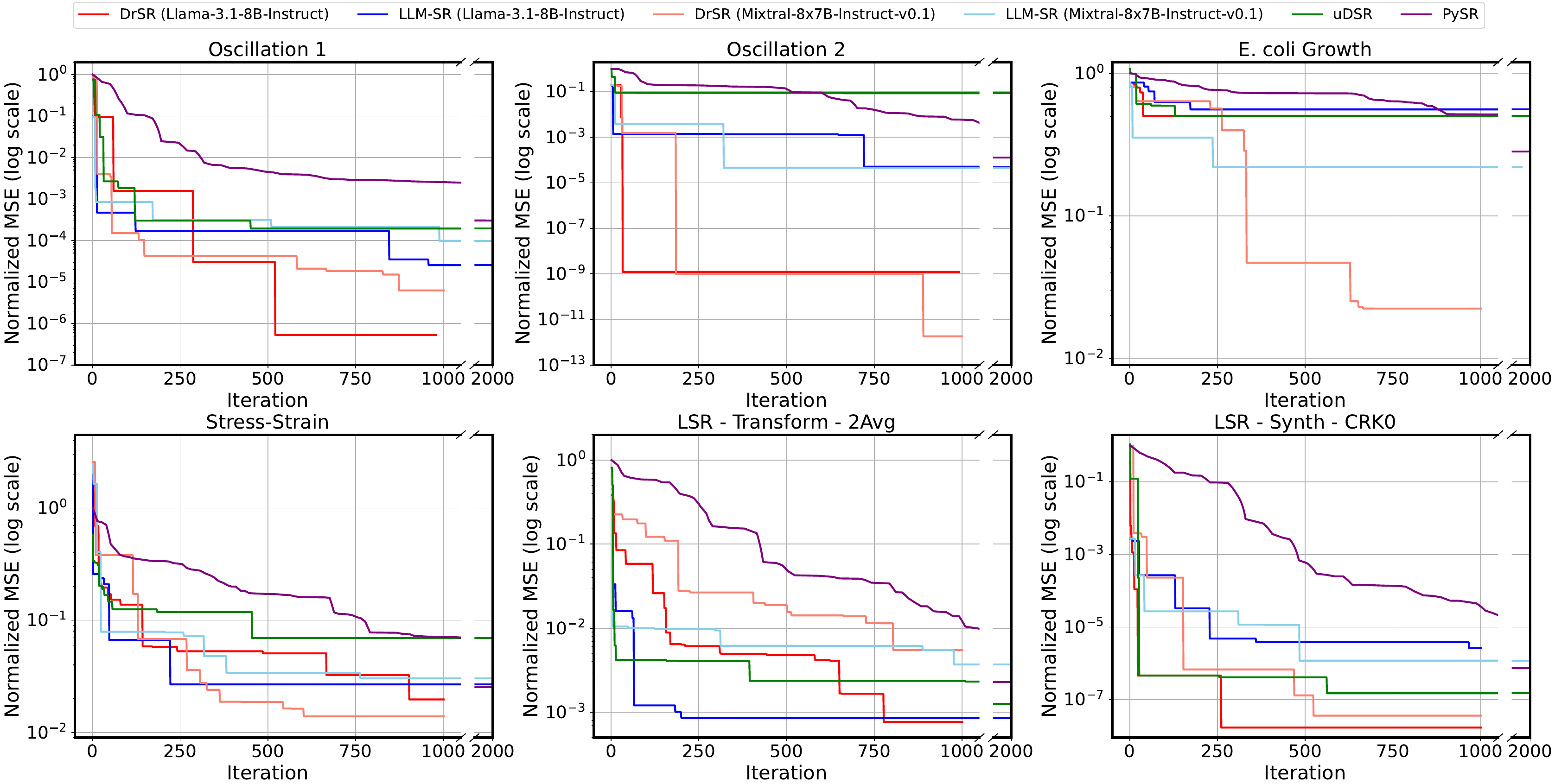}  
  \caption{
  \textbf{Training convergence comparison.}
  DrSR demonstrates faster convergence and lower final error across all tasks, consistently outperforming baselines even in early-stage iterations. 
  }
  \label{fig:convergence}
\end{figure}

We evaluate the convergence efficiency of DrSR relative to LLM-SR, uDSR, and PySR on six benchmark tasks. Figure~\ref{fig:convergence} shows the NMSE over training iterations. DrSR converges significantly faster than all baselines and achieves lower final errors. In most cases, it matches or surpasses the $2000$-iteration performance of others within just $1000$ steps. 
This efficiency stems from DrSR's dual-loop reasoning: data-driven insight narrows the search space, while experience-guided generation avoids redundant or suboptimal candidates. Together, they enable more effective optimization under limited iteration budgets.

\subsection{Valid Solution Rate Analysis}

\begin{figure}[H]
  \centering
  \includegraphics[width=0.9\textwidth]{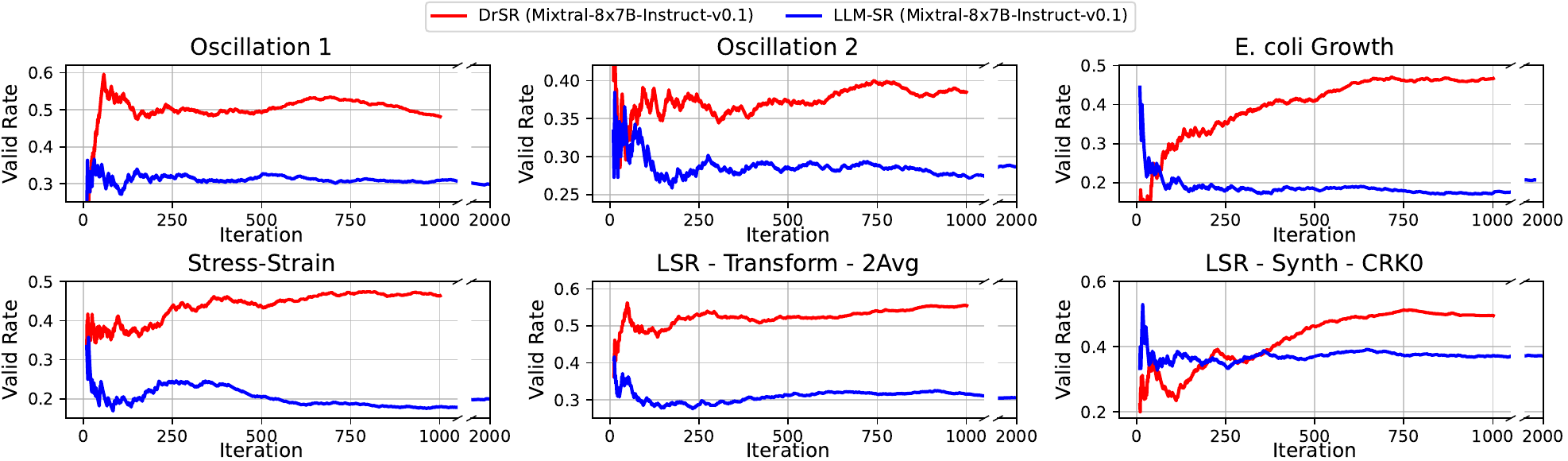}  
  \caption{
  \textbf{Valid solution rate comparison.}
  We report the proportion of syntactically valid, compilable, and evaluable equations produced by DrSR and LLM-SR.
  }
  \label{fig:validrate}
\end{figure}

To further explain the superior efficiency of DrSR, we evaluate the \textit{valid solution rate}—the proportion of generated equations that are syntactically valid, compilable, and executable. Figure~\ref{fig:validrate} compares this rate for DrSR and LLM-SR (both using Mixtral). DrSR consistently yields a higher valid rate across all tasks, especially on structurally complex datasets like E. coli Growth and Stress-Strain, where it maintains 0.4--0.6 validity compared to LLM-SR’s sub-0.3 plateau. This improvement is largely due to DrSR’s Idea Library, which captures and applies error-avoidance strategies extracted from prior failures. By reducing the frequency of invalid generations, DrSR lowers computational waste and improves sample efficiency—key advantages in practical SR under limited resources.

\subsection{Ablation Study}

We conduct ablation experiments on the Oscillator 2 task using Mixtral backbone to evaluate the contribution of DrSR’s two key components: \textit{Data-aware Insight} and \textit{Inductive Idea Extraction}.

\begin{figure}[t]
  \centering
  \includegraphics[width=0.65\textwidth]{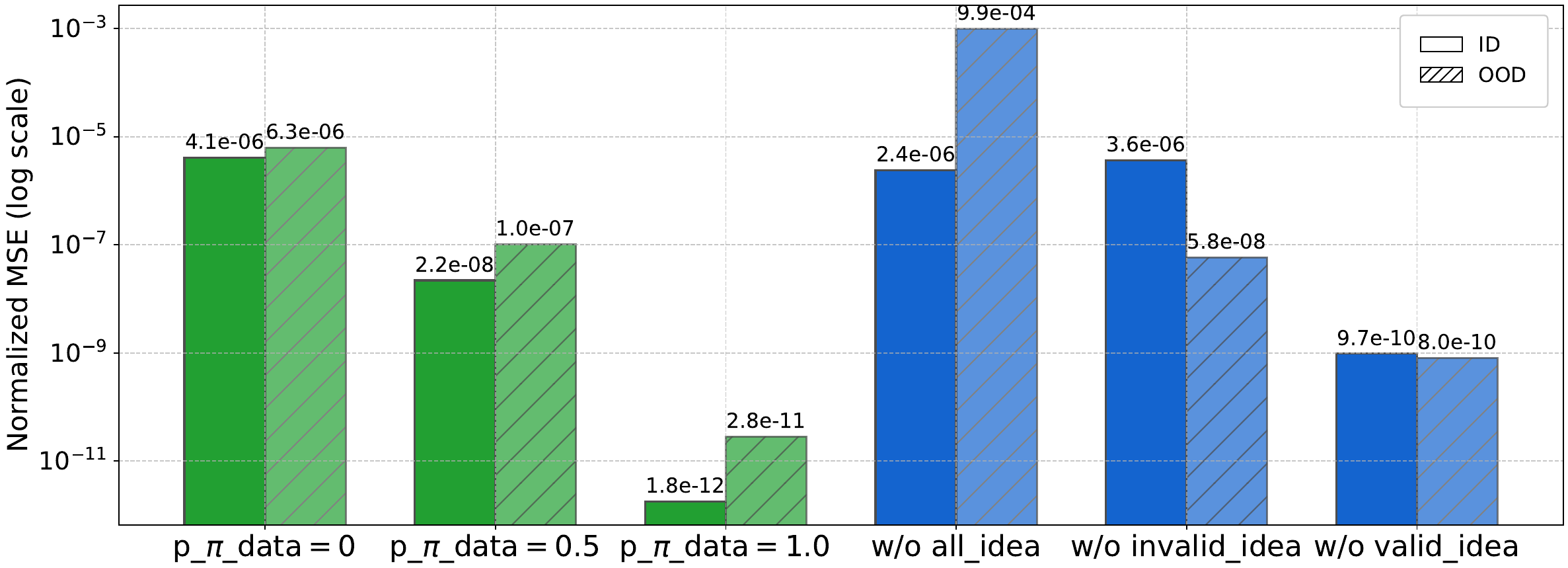}  
  \caption{
  \textbf{Ablation results on the Oscillation 2 problem, }
  showing the impact of Data-aware Insight and Inductive Idea Extraction on the performance of DrSR. 
  }
  \label{fig:ablation}
  \vspace{-15pt}
\end{figure}

\textbf{Impact of Data-aware Insight.}
We vary the probability \( p \in \{0, 0.5, 1.0\} \) with which the main LLM references structured insights during generation. Setting \( p=0 \) disables data priors, while \( p=1.0 \) enforces full use. Performance improves consistently with higher usage of insights. Without data priors, the model exhibits LLM-SR-like behavior and inefficient search. In contrast, full integration (\( p=1.0 \)) significantly accelerates convergence and improves accuracy, confirming that structural guidance sharpens the search space and enhances model stability.

\vspace{-0.2mm}

\textbf{Effect of Inductive Idea Extraction.}
To assess the importance of inductive feedback, we incrementally disable parts of the Inductive Idea Extraction module while keeping data insights active:
\textit{w/o valid ideas} removes strategies from successful equations; 
\textit{w/o invalid ideas} also removes error-avoidance heuristics; 
\textit{w/o all ideas} disables the module entirely, reducing DrSR to LLM-SR. Removing any component degrades performance, with losses compounding across settings. Valid ideas guide structural preference, while invalid ideas prevent repeat errors. Their combination enables targeted exploration, proving critical to DrSR’s performance.

\vspace{-1.2mm}

\section{Related Work}
\label{headings}

\vspace{-1.2mm}

\textbf{Classical Symbolic Regression.}
SR has long served as a core tool in scientific modeling~\cite{gerwin1974information, langley1977bacon}, with growing relevance in the AI4Science community~\cite{makke2024interpretable, merler2024context}. Early methods explored expression discovery via linear modeling, GP, evolutionary search, and reinforcement learning~\cite{koza1994genetic, schmidt2009distilling, cranmer2023interpretable, petersen2019deep, sun2022symbolic}. Sparse regression techniques~\cite{brunton2016discovering, champion2019data} and symbolic systems modeling~\cite{martius2016extrapolation, sahoo2018learning, kim2020integration, wu2024prunesymnet, li2025metasymnet} have extended SR to complex, nonlinear domains. GP-based approaches, initiated by Koza~\cite{koza1990genetic}, continue to evolve with modern variants~\cite{de2021interaction, he2022taylor, zhang2023sr}, while RL-based methods~\cite{petersen2019deep, mundhenk2021symbolic,landajuela2021improving, crochepierre2022interactive, li2023neural, du2024discover, pettit2025disco} offer adaptive equation construction under search constraints. Recently, Transformer-based models such as SymbolicGPT and SymFormer~\cite{biggio2021neural,valipour2021symbolicgpt, kamienny2022end, vastl2024symformer, li2022transformer, yu2024symbolic,ding2024niert} have introduced large-scale pretraining to symbolic tasks. Despite their strong generative capabilities, these models often lack the ability to incorporate prior scientific knowledge, which is a key advantage of LLMs.

\vspace{-0.2mm}

\textbf{LLMs for Symbolic Regression.}
LLMs have spurred a new paradigm in SR by enabling language-driven equation generation~\cite{shojaee2024llm, grayeli2024symbolic, merler2024context, meyerson2024language, guo2024coevo}. LLM-SR~\cite{shojaee2024llm} employs LLMs to produce equation skeletons guided by prior scientific knowledge, followed by parameter optimization. LaSR~\cite{grayeli2024symbolic} extends this by incorporating abstract symbolic patterns; CoEvo~\cite{guo2024coevo} leverages open-ended knowledge co-evolution during search. ICSR~\cite{merler2024context} encodes training data as in-context demonstrations to generate candidate expressions. In contrast to these approaches, DrSR introduces a closed-loop reasoning framework that augments LLMs with structural insights and experience-driven reflection. By explicitly modeling the interaction between observation, generation, and feedback, DrSR enables more efficient and generalizable symbolic discovery beyond existing LLM-based variants.

\vspace{-1.2mm}

\section{Conclusion and Future Directions}

\vspace{-1.2mm}

We propose DrSR, a framework that augments LLM with data-driven insight and experience-guided reflection, enabling both interpretation of input data and learning from past generations. This dual reasoning design overcomes key limitations of prior LLM-based methods and delivers superior performance across diverse scientific problems, outperforming both traditional and LLM-based baselines in accuracy, generalization, and convergence. Despite its strong empirical results, DrSR also opens promising avenues for future research, including multimodal extensions to accommodate richer data types such as scientific imagery, and continual learning to accumulate transferable modeling strategies across tasks. These directions represent exciting opportunities for future exploration, and we plan to pursue them to further advance DrSR toward next-generation scientific equation discovery.

\newpage
\bibliographystyle{unsrt}
\bibliography{ref}

\newpage

\section*{Appendix}

\subsection*{A: Implementation and Experimental Details}
\label{appendix:implementation}

\subsubsection*{A.1 Baseline Configurations}

For \texttt{gplearn}, we use a symbolic regressor with a population size of 1000, tournament size of 20, and 2000 generations. The function set includes standard arithmetic and common transcendental operators (\texttt{add}, \texttt{sub}, \texttt{mul}, \texttt{div}, \texttt{sin}, \texttt{cos}, \texttt{sqrt}, \texttt{log}, \texttt{abs}, \texttt{neg}, \texttt{inv}). The fitness metric is mean squared error (MSE), with a parsimony coefficient of 0.01 to encourage simpler expressions. For \texttt{PySR} (version 0.18.2), the model is configured with a population size of 50, tournament selection size of 10, and 2000 iterations. We use standard binary operators (\texttt{+}, \texttt{-}, \texttt{*}, \texttt{/}) and common unaries such as \texttt{sin}, \texttt{cos}, \texttt{sqrt}, \texttt{log}, and \texttt{abs}. For \texttt{DSR}, we adopt the default configuration with a batch size of 100, number of samples set to 200{,}000, and training noise parameter \texttt{epsilon} = 0.05. The symbolic operator set includes \{\texttt{add}, \texttt{sub}, \texttt{mul}, \texttt{div}, \texttt{sin}, \texttt{cos}, \texttt{exp}, \texttt{log}\}, and the reward metric is negative normalized MSE (\texttt{neg-NMSE}). We also apply entropy-based policy regularization during optimization. The \texttt{uDSR} configuration builds upon \texttt{DSR} with several extensions: the function set includes additional symbolic components such as \texttt{poly} and \texttt{const}; the prior length is capped at 20; and a polynomial optimizer is employed with degree 2 and coefficient tolerance of \(1 \times 10^{-5}\). The training settings (n\_samples = 200{,}000, batch\_size = 100, \texttt{epsilon} = 0.05) match those used in \texttt{DSR}, with separate entropy and learning rate parameters adjusted for stability. For LLM-based baselines such as \texttt{LLM-SR} and \texttt{LaSR}, we use the official configurations without additional tuning or prompt modifications.

\subsubsection*{A.2 DrSR Configuration}

DrSR retains the symbolic skeleton generation structure of LLM-SR while incorporating two additional reasoning modules. For \textit{Data-aware Insight} (\(\pi_{\text{data}}\)), 100 input-output pairs are sampled to estimate data structure (e.g., monotonicity, correlation), refreshed after every performance improvement. Sampling uses temperature 0.6, top-\(k = 30\), and top-\(p = 0.3\). For \textit{Inductive Idea Extraction} (\(\pi_{\text{idea}}\)), ideas are extracted based on categorized evaluation outcomes. At each iteration, up to three ideas are drawn from the last 50\% of the existing entries per category (positive, negative, error), stored in JSON format, and reused in prompting.

\subsubsection*{A.3 Large Language Models}

DrSR uses two backbone LLMs: \texttt{LLaMA-3.1-8B-Instruct} and \texttt{Mixtral-8x7B-Instruct-v0.1}, both quantized and deployed on NVIDIA H100 GPUs. Notably, \texttt{LLaMA-3.1-8B} requires only around 8GB of VRAM in 4-bit quantized form, suggesting that DrSR can also be run on less powerful machines with proper quantization. We use the same decoding parameters as LLM-SR and optimize four sampled candidate equations per iteration using BFGS. Each DrSR experiment runs for a maximum of 1000 iterations. On our setup, it takes roughly 6 hours to complete on \texttt{LLaMA-3.1-8B-Instruct} and 12 hours on \texttt{Mixtral-8x7B-Instruct-v0.1}.

\subsection*{B: Dataset Descriptions}
\label{appendix:datasets}

To evaluate the generality and effectiveness of DrSR, we conduct experiments on seven benchmark tasks drawn from two major sources: datasets used in the original LLM-SR study and the LLM-SRBench suite. These datasets span a variety of scientific domains, including physics, biology, chemistry, and materials science. Each dataset is associated with an underlying ground-truth expression, which serves as the target for symbolic discovery. The descriptions in Sections B.1 through B.3 are adapted from the dataset documentation in LLM-SR.

\subsubsection*{B.1 Nonlinear Oscillators}
Nonlinear damped oscillators represent a fundamental class of systems in physics and engineering, where the motion of a mass under restoring and damping forces is modeled by second-order differential equations. These systems capture the interplay between displacement, velocity, and complex external forces. Mathematically, the dynamics follow the form \(\ddot{x} + f(t, x, \dot{x}) = 0\), where \(f\) denotes a nonlinear force term dependent on time, position, and velocity.

LLM-SR simulates two custom oscillator systems to evaluate model robustness in learning nontrivial physical relationships. Oscillator 1 is defined by:
\[
\dot{v} = F \sin(\omega x) - \alpha v^3 - \beta x^3 - \gamma x v - x \cos(x)
\hspace{1em}
\scriptsize
(F = 0.8,\ \alpha = 0.5,\ \beta = 0.2,\ \gamma = 0.5,\ \omega = 1.0)
\]
and Oscillator 2 is defined by:
\[
\dot{v} = F \sin(\omega t) - \alpha v^3 - \beta x v - \delta x \exp(\gamma x)
\hspace{1em}
\scriptsize
(F = 0.3,\ \alpha = 0.5, \ \beta = 1.0, \ \delta = 5.0, \ \gamma = 0.5,\ \omega = 1.0)
\]
with \(v = \dot{x}\). Both are initialized with \(x = 0.5\), \(v = 0.5\), and simulated over \(t \in [0, 50]\). These formulations introduce strong nonlinearities and variable interactions to prevent trivial memorization and encourage structural reasoning.

 These systems were chosen to highlight symbolic regression models’ ability to generalize under different nonlinear dynamics and to benchmark performance beyond conventional oscillatory systems.

\subsubsection*{B.2 Bacterial Growth}
Modeling the growth dynamics of Escherichia coli plays a vital role in various scientific and engineering disciplines, including microbiology, biotechnology, and food safety. Accurate symbolic models that describe E. coli population expansion under diverse environmental conditions are critical for both scientific understanding and real-world applications. 

To evaluate symbolic regression models in such contexts, LLM-SR adopt a biologically grounded yet nontrivial benchmark formulation. The population growth rate is governed by a nonlinear differential equation that incorporates four major environmental factors: population density \(B\), substrate concentration \(S\), temperature \(T\), and pH level. The general form is:
\[
\frac{dB}{dt} = f(B, S, T, \text{pH}) = f_B(B) \cdot f_S(S) \cdot f_T(T) \cdot f_{\text{pH}}(\text{pH})
\]
where each multiplicative term controls a different aspect of bacterial behavior. To increase modeling difficulty and avoid trivial memorization, we introduce customized nonlinear designs for \(f_T(T)\) and \(f_{\text{pH}}(\text{pH})\) while maintaining realistic biological interpretations.

The full form of the growth equation used in our benchmark is:
\[
\frac{dB}{dt} = \mu_{\text{max}} B \left( \frac{S}{K_S + S} \right)
\left(\frac{\tanh{k(T - x_0)}}{1 + c(T - x_{\text{decay}})^4}\right)
\exp\left(-|\text{pH} - \text{pH}_{\text{opt}}|\right)
\sin\left( \frac{(\text{pH} - \text{pH}_{\text{min}})\pi}{\text{pH}_{\text{max}} - \text{pH}_{\text{min}}} \right)^2
\]

This dataset presents a comprehensive challenge for symbolic solvers by integrating multiple biological priors into a structurally complex formulation. The design encourages models not only to capture meaningful interactions, but also to generalize beyond simplistic template recall.

\subsubsection*{B.3 Material Stress Behavior}

The stress-strain characteristics of materials under varying thermal conditions are fundamental to structural design and materials engineering. In this benchmark, LLM-SR adopts a real-world experimental dataset, capturing the tensile behavior of Aluminum 6061-T651 under uniaxial tension across six distinct temperatures, ranging from 20\textdegree C to 300\textdegree C. This dataset provides a practical and non-synthetic scenario, presenting significant challenges for symbolic regression models.

Unlike traditional physics-based benchmarks with known governing equations, this task does not follow a predefined theoretical model. Instead, it demands empirical discovery of complex, temperature-dependent relationships—highlighting the need for expressive modeling capacity and strong generalization ability. Notably, the symbolic expression must be inferred purely from observation without memorizing known formulas, pushing LLM-based frameworks beyond their reliance on learned priors.

To describe the temperature-dependent stress-strain relationship, one commonly used empirical formulation is:
\[
\sigma = \left( A + B \varepsilon^n \right) \left( 1 - \left( \frac{T - T_r}{T_m - T_r} \right)^m \right)
\]
where $\sigma$ is the stress, $\varepsilon$ is the strain, $T$ is the temperature, $T_r$ and $T_m$ denote reference and melting temperatures, respectively, and $A$, $B$, $n$, and $m$ are material-specific parameters. 

\subsubsection*{B.4 LSR-Transform}

The LSR-Transform dataset, as introduced in LLM-SRBench, is constructed by applying symbolic transformations to problems in the Feynman benchmark. These transformations involve solving for alternative target variables and reformatting equations into mathematically valid but less familiar forms. This design prevents trivial memorization by large language models (LLMs) and emphasizes scientific reasoning and data-driven discovery.

In our DRSR experiments, we specifically selected two tasks from LSR-Transform:

\paragraph{I.37.4\_0\_1} This problem involves discovering the expression for the resultant wave intensity $I_1$ in terms of phase difference $\delta$, and intensities $I_2$ and $Int$ from two wave sources. The transformed ground-truth equation is:
\[
I_1 = 2 I_2 \cos^2(\delta) + I_2 + Int + 2 \sqrt{I_2 \left( I_2 \cos^2(\delta) + I_2 + Int \right)} \cos(\delta)
\]
Here, $I_1$ represents the resultant intensity, $I_2$ is the intensity of the second wave source, $Int$ is the intensity of the first source, and $\delta$ is the phase difference.

\paragraph{III.4.33\_3\_0} This problem originates from quantum harmonic oscillator systems. The target equation, with temperature $T$ as the dependent variable, is:
\[
T = \frac{h \omega}{2 \pi k_b \log\left(1 + \frac{h \omega}{2 \pi E_n}\right)}
\]
In this context, $T$ is the system temperature, $E_n$ is the energy of the $n$th mode, $h$ is Planck's constant, $\omega$ is the angular frequency, and $k_b$ is Boltzmann’s constant.

These transformed tasks test the ability of LLM-based methods to reason through uncommon representations of well-known physical principles.

\subsubsection*{B.5 LSR-Synth}

The LSR-Synth dataset is designed to evaluate models' ability to discover novel equations that combine standard scientific components with synthetic terms not typically found in literature. These problems span four domains—chemistry, physics, biology, and materials science—and are validated for solvability and plausibility by human experts.

In our experiments, we included the chemistry task \textbf{CRK0}, which models reaction kinetics. The target differential equation is:
\[
\frac{dA}{dt} = -0.1899 \cdot A(t)^2 + \frac{0.1899\_z \cdot A(t)^2}{0.7498 \cdot A(t)^4 + 1}
\]
where $A(t)$ denotes concentration at time $t$, and $dA/dt$ is the reaction rate. This equation incorporates a second-order decay term along with a synthetic nonlinear saturation term, making it an ideal test case for evaluating symbolic reasoning in data-driven kinetic modeling.

\subsection*{C: Prompt}

\begin{figure}[htbp]
  \centering
  \includegraphics[width=\linewidth]{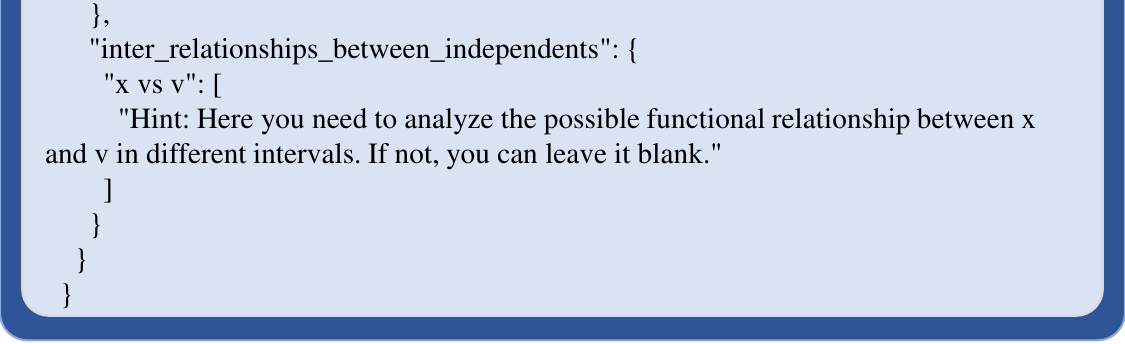}
  \caption{Prompt for Data-aware Insight.}
  \label{fig:app_prompt_1}
\end{figure}

The prompts used to generate Data-aware Insights for Oscillators 1 are shown in Fig~\ref{fig:app_prompt_1}, where Prompt 1 corresponds to the prompt for Initial Data-aware Insight, Prompt 2 corresponds to the prompt used in Iterative Residual-based Refinement, and Prompt 3 shows the shared task requirements for Prompt 1 and 2.

\begin{figure}[htbp]
  \centering
  \includegraphics[width=\linewidth]{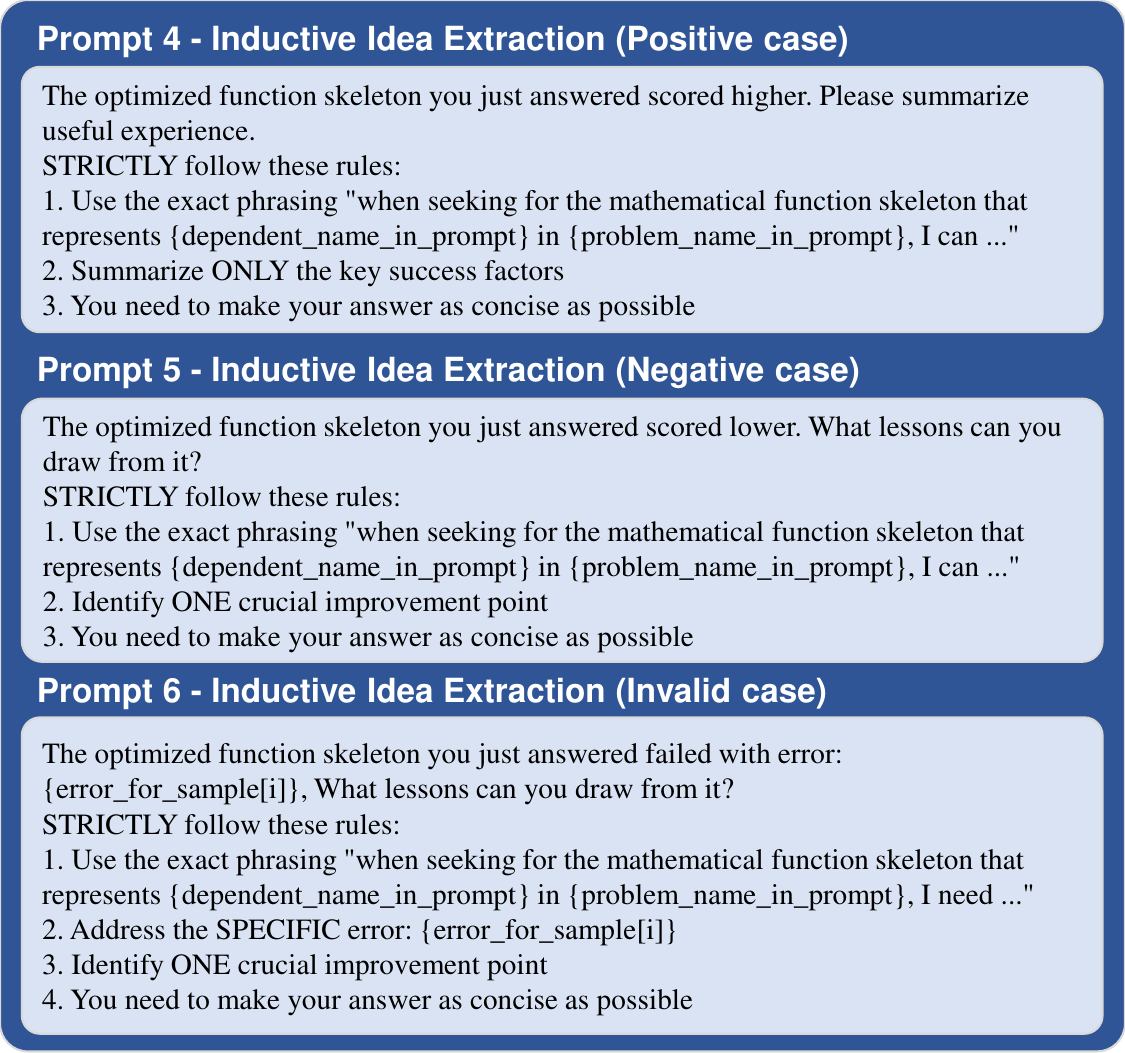}
  \caption{Prompt for Inductive Idea Extraction.}
  \label{fig:app_prompt_2}
\end{figure}

\noindent The prompts used in Inductive Idea Extraction for Oscillators 1 are shown in Fig~\ref{fig:app_prompt_2}, corresponding respectively to the cases where the equation under analysis is Positive, Negative, and the output is Invalid.

\begin{figure}[htbp]
  \centering
  \includegraphics[width=\linewidth]{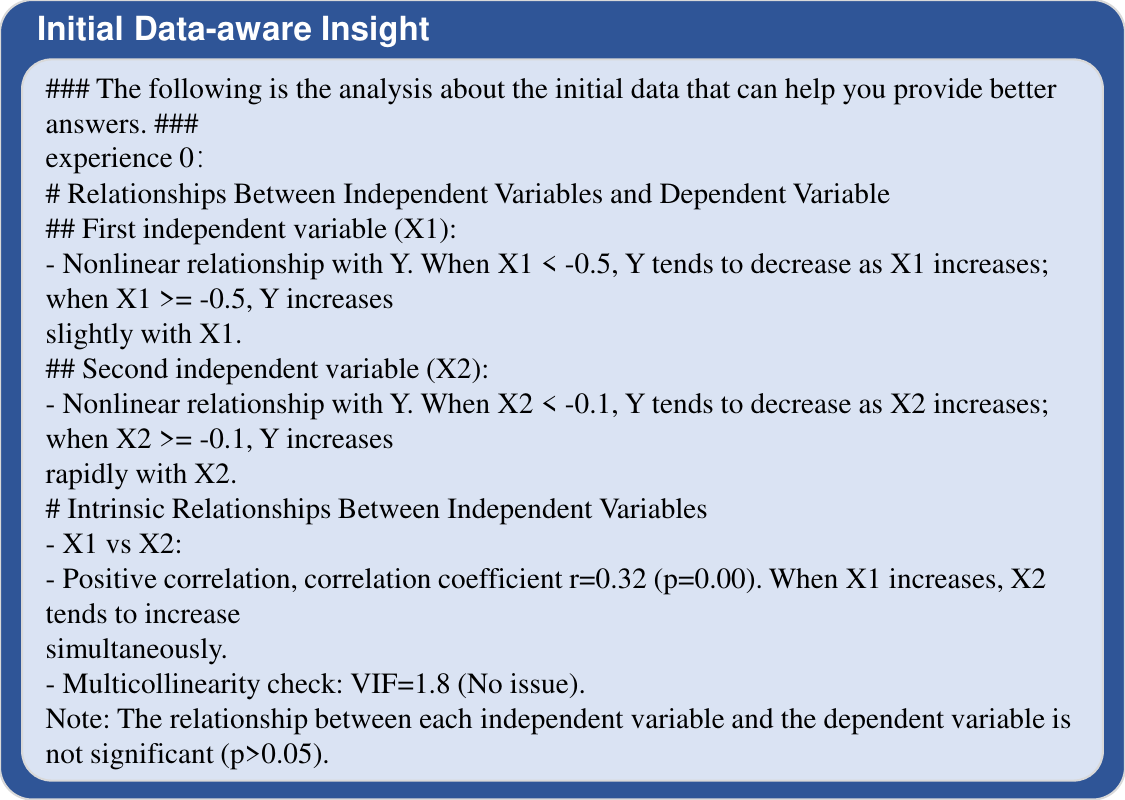}
  \caption{Initial Data-aware Insight.}
  \label{fig:app_prompt_res0}
\end{figure}

\begin{figure}[htbp]
  \centering
  \includegraphics[width=\linewidth]{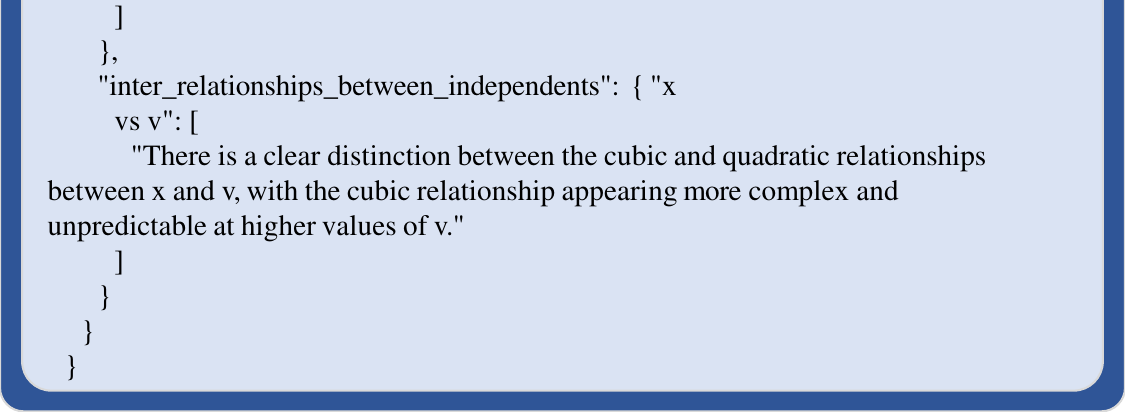}
  \caption{Final Data-aware Insight.}
  \label{fig:app_prompt_res1}
\end{figure}

\begin{figure}[htbp]
    \centering
    \includegraphics[width=\linewidth]{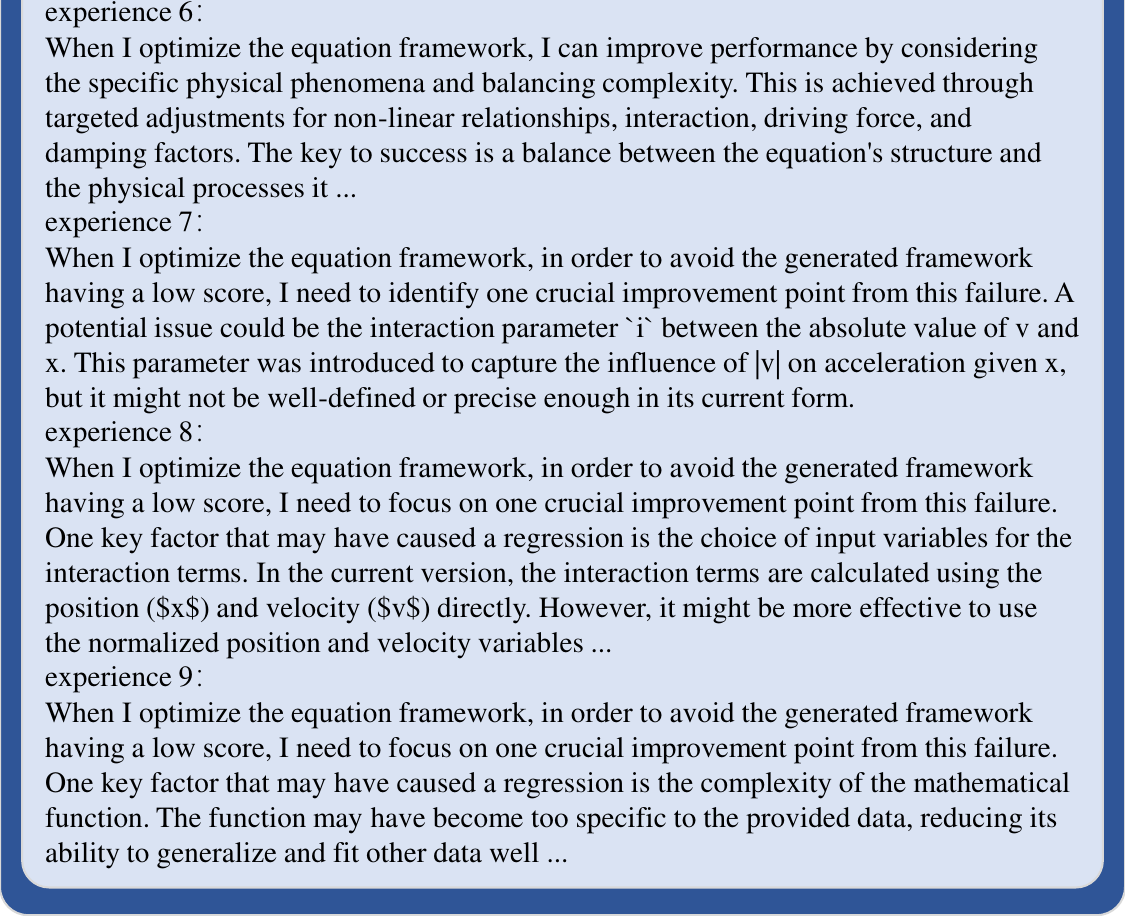}
    \caption{Final Inductive Idea Extraction.}
    \label{fig:app_prompt_idea}
\end{figure}

\noindent Fig~\ref{fig:app_prompt_res0}, \ref{fig:app_prompt_res1} and \ref{fig:app_prompt_idea} depict the evolution of our prompt-guided components, including \textit{Data-aware Insight} and \textit{Inductive Idea Extraction}, throughout the discovery process. Fig~\ref{fig:app_prompt_res0} corresponds to the initial insight generated at the beginning of the search. Fig~\ref{fig:app_prompt_res1} and \ref{fig:app_prompt_idea} present the insights and ideas extracted when the model discovers its best-performing equation. These visualizations indicate that the model's \textit{Data-aware Insight} of the dataset becomes progressively deeper, and that \textit{Inductive Idea Extraction} are gradually refined and enriched over time.

\begin{figure}[htbp]
  \centering
  \includegraphics[width=\linewidth]{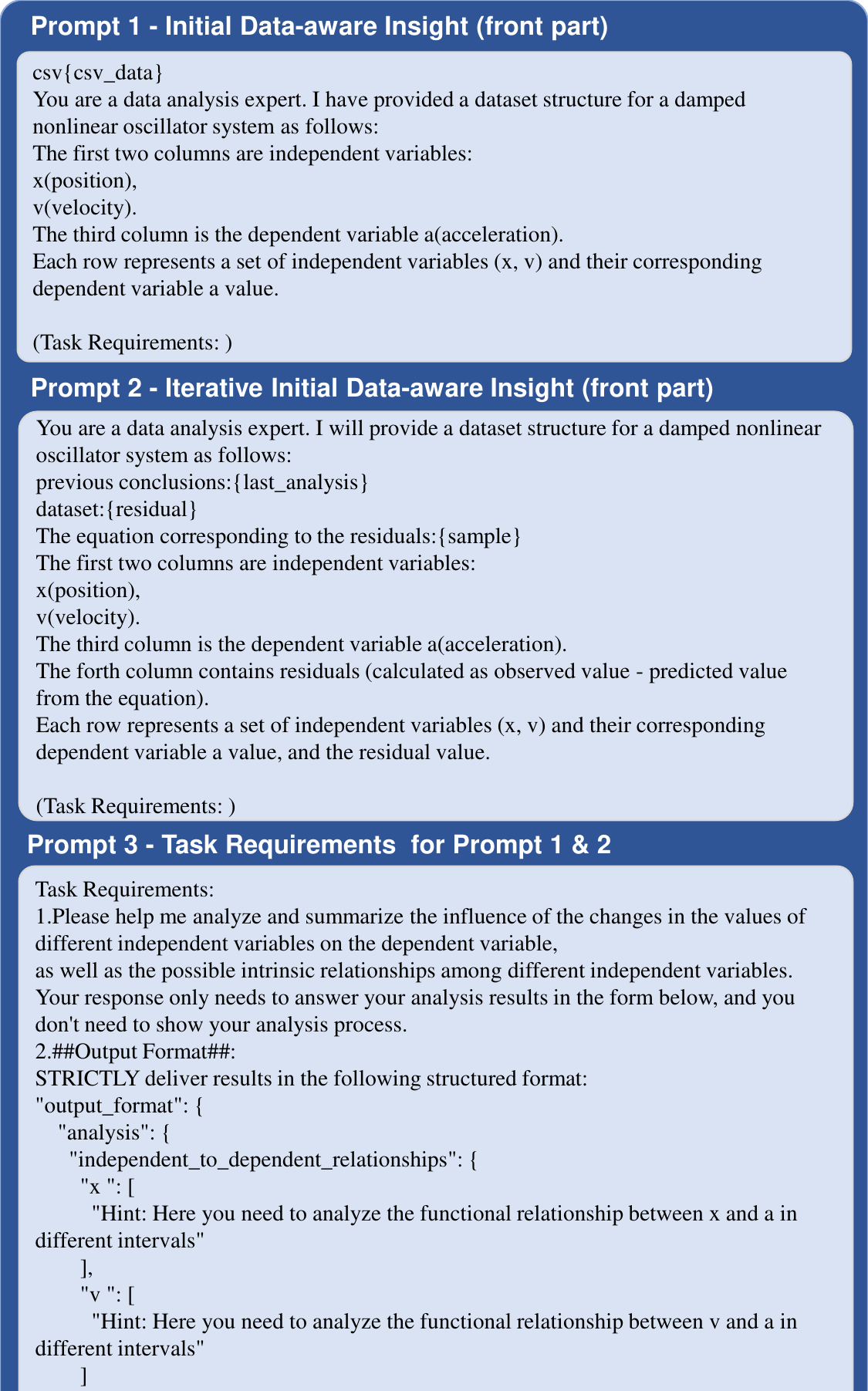}
\end{figure}

\begin{figure}[htbp]
  \centering
  \includegraphics[width=\linewidth]{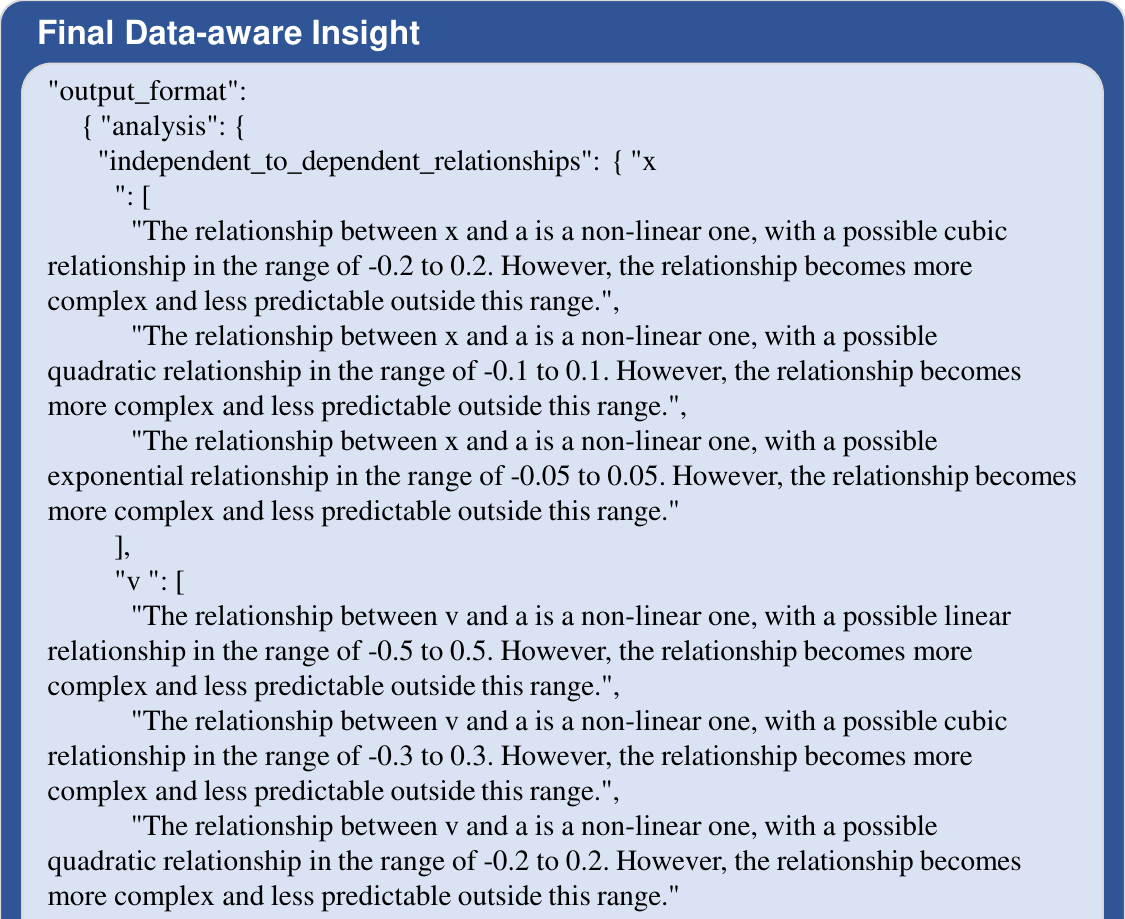}
\end{figure}

\begin{figure}[htbp]
    \centering
    \includegraphics[width=\linewidth]{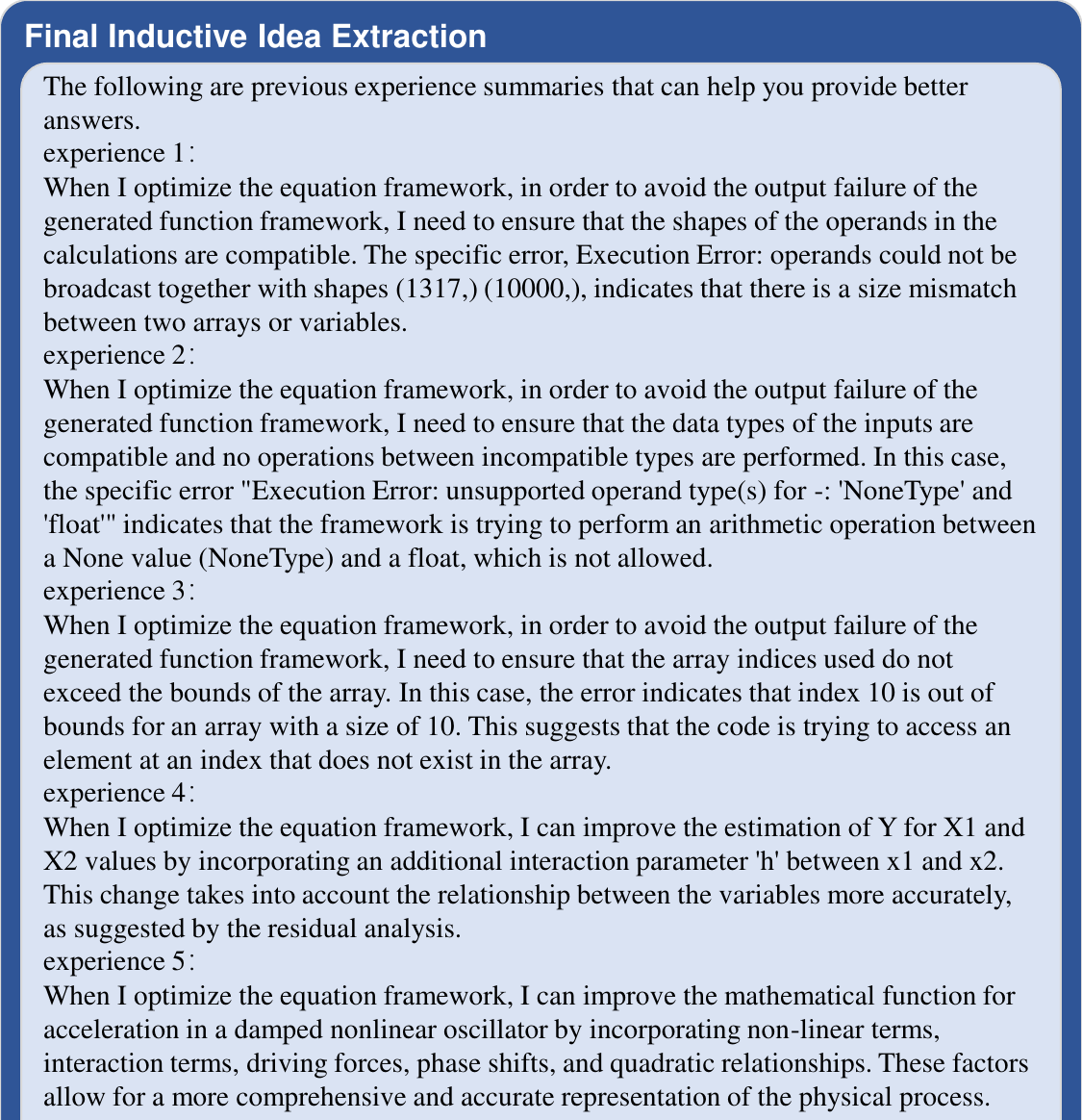}
\end{figure}

\subsection*{D: Results Presentation}

\begin{figure}[h]
    \centering
    \includegraphics[width=0.8\textwidth]{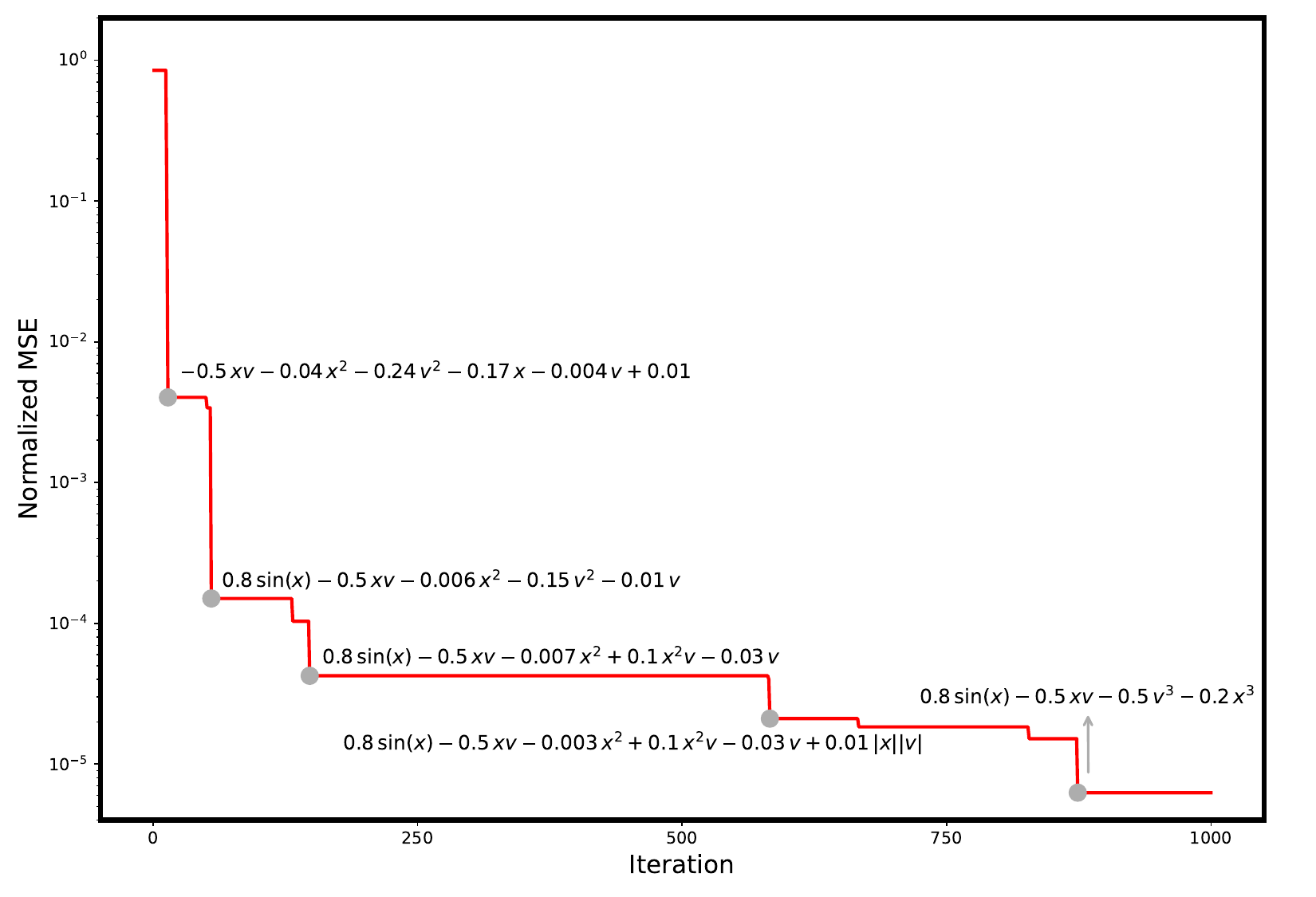}
    \caption{Performance trajectory of DrSR with the best-performing equation programs discovered over iterations on the Oscillation 1 problem.}
    \label{fig:app_drsr_trajectory}
\end{figure}

\noindent Fig~\ref{fig:app_drsr_trajectory} illustrates the normalized mean squared error (NMSE) values on the Oscillation 1 dataset. For clarity, we present the discovered equation programs in their simplified mathematical form, accompanied by optimized parameters. The program initially explores polynomial terms; as iterations progress and data understanding improves, irrelevant polynomial and constant terms are gradually eliminated. Notably, nonlinear terms—such as sinusoidal components—are identified early and consistently preserved throughout the search.

\noindent The ground-truth equation for this dataset is:
\[
\dot{v} = 0.8 \sin(x) - 0.5 x\cdot v - 0.5v^3 - 0.2x^3 - x \cos(x),
\]
the best equation discovered after 1000 iterations by DrSR is:
\[
\dot{v} = 
\colorbox{pink}{$0.8 \sin(x)$}
- \colorbox{pink}{$0.5 x\cdot v$}
- \colorbox{pink}{$0.5 v^3$}
- \colorbox{pink}{$0.2 x^3$}
\]
whereas the best equation discovered after 2000 iterations by LLM-SR is:
\[
\dot{v} =
-0.2 x
- 0.02 v
+ 6 \times 10^{-6} x^2
- 0.03 v^2
- \colorbox{pink}{$0.5 x\cdot v$}
+ 0.08 \colorbox{pink}{$v^3$}
+ 0.15 \colorbox{pink}{$x^3$}
\]

The terms highlighted in pink indicate the exact components of the ground-truth equation that are correctly predicted by DrSR. This comparison highlights that DrSR captures core structural terms of the ground-truth equation, showing significantly closer alignment than LLM-SR. We attribute this improvement to DrSR’s ability to integrate data-driven insight and reflective feedback in the discovery process.

\subsection*{E: Limitation}
\label{Limitation}

\textbf{Limitations.} While DrSR shows strong potential for enabling LLM-guided scientific discovery, it also has several limitations. First, due to the inherent stochasticity of large language models, the generated  outputs can occasionally be verbose, repetitive, or overly complex, requiring expert intervention to interpret or refine. Second, the parameter optimization phase in DRSR currently relies on the BFGS algorithm, which, while efficient, may not always achieve globally optimal parameter values for complex equation skeletons. Exploring more robust or adaptive optimization strategies remains an important direction for future work.

\subsection*{F: Gaussian Noise Evaluation}
To simulate the task of DrSR in real-world environments, we conducted a systematic analysis of the model's performance under different noise levels to evaluate its robustness. We performed controlled experiments by introducing Gaussian noise with different standard deviations ($\sigma = \{0.001, 0.002\}$) into the training dataset of Oscillation 1, and observed the NMSE and $\text{ACC}_\tau$ performance on the test sets (ID/OOD).

\begin{table}[h]
\centering
\setlength{\tabcolsep}{4pt}  
\caption{Comparison of DrSR and LLM-SR on Oscillator1 under different Gaussian noise levels ($\sigma = \{0.001, 0.002\}$), evaluated with NMSE and accuracy under ID and OOD settings. All results use LLaMA3.1 as backbone.}
\begin{tabular}{l*{8}{>{\centering\arraybackslash}p{0.09\linewidth}}}
\toprule
\multirow{3}{*}{Model} 
& \multicolumn{4}{c}{\(\sigma = 0.001\)} 
& \multicolumn{4}{c}{\(\sigma = 0.002\)} \\
\cmidrule(lr){2-5} \cmidrule(lr){6-9}
& \multicolumn{2}{c}{ID} & \multicolumn{2}{c}{OOD}
& \multicolumn{2}{c}{ID} & \multicolumn{2}{c}{OOD} \\
\cmidrule(lr){2-3} \cmidrule(lr){4-5} \cmidrule(lr){6-7} \cmidrule(lr){8-9}
& NMSE\(\downarrow\) & Acc\(_{0.01}\)\(\uparrow\) 
& NMSE\(\downarrow\) & Acc\(_{0.01}\)\(\uparrow\) 
& NMSE\(\downarrow\) & Acc\(_{0.01}\)\(\uparrow\) 
& NMSE\(\downarrow\) & Acc\(_{0.01}\)\(\uparrow\) \\
\midrule
\textbf{LLM-SR} & \(1.59e-5\) & \(89.20\%\) & \(0.0021\) & \(10.46\%\) & \(2.68e-5\) & \(82.08\%\) & \(0.0025\) & \(10.50\%\) \\
\textbf{DrSR}   & \(6.46e-6\) & \(93.24\%\) & \(0.0006\) & \(31.98\%\) & \(2.03e-6\) & \(95.39\%\) & \(0.0011\) & \(14.46\%\) \\
\bottomrule
\end{tabular}
\label{tab:oscillator1_noise}
\end{table}

Table~\ref{tab:oscillator1_noise} presents a comparative analysis of DrSR and LLM-SR under different noise conditions. The results show that as the noise level increases, both models demonstrate noise resistance to data fluctuations. However, compared to LLM-SR, DrSR consistently achieves significantly better NMSE and $\text{ACC}_\tau$ scores.


\end{document}